# Two-Stage Grasping: A New Bin Picking Framework for Small Objects

Hanwen Cao, *Student Member*, *IEEE*, Jianshu Zhou, *Member*, *IEEE*, Rui Cao, Junda Huang,
Yichuan Li, Ng Cheng Meng, Qi Dou, and Yunhui Liu, *Fellow*, *IEEE*

*Abstract*—This paper proposes a novel bin picking framework, two-stage grasping, aiming at precise grasping of cluttered small objects. Object density estimation and rough grasping are conducted in the first stage. Fine segmentation, detection, grasping, and pushing are performed in the second stage. A small object bin picking system has been realized to exhibit the concept of two-stage grasping. Experiments have shown the effectiveness of the proposed framework. Unlike traditional bin picking methods focusing on vision-based grasping planning using classic frameworks, the challenges of picking cluttered small objects can be solved by the proposed new framework with simple vision detection and planning.

## I. Introduction

Bin picking is a canonical problem in robotics. It aims to pick identical objects one by one from a cluttered bin [1]. In recent years, bin picking has drawn considerable attention due to its wide applications in modern logistics [2, 3] and the service industry [4, 5]. Though its importance and ubiquity, bin picking remains challenging.

Recent bin picking solutions use learning-based image processing techniques, such as image segmentation and pose estimation, to detect grasps that can pick one item out of the bins. For example, [2] used a semantic segmentation system and a heuristic suction grasp generation algorithm to pick specific items from a cluttered pile of objects. The grasp performance is related to semantic segmentation quality, and the suction grasp quality is influenced by point cloud resolution and precision. [6, 7] introduced reflective object bin picking methods and datasets. These methods are based on scene reconstruction and 6D object pose estimation. [8, 9] introduced vision-based object recognition and pose regression solutions in bin picking scenarios. These approaches have achieved remarkable results in picking normal-sized industrial items [7] and everyday household objects [10]. However, a large variety of objects remain difficult for robots. Among them, small object bin picking is especially challenging for the following reasons. Firstly, sensing cluttered small objects is challenging. The state-of-the-art image segmentation approaches still have difficulty distinguishing image pixels belonging to different small instances [11, 12]. Object pose estimation methods require precise sensing of complete point clouds or depths [13]. Both small object sizes and complex object shapes make sensor resolutions and occlusion limitations to pose estimation [14], [15]. Furthermore, many small objects, such as metal nuts and bolts, have reflective surfaces, which causes unpredictable sensing results for commercially available 3D cameras [16]. Secondly, the existing bin picking approaches rely on a widely accepted one-stage (sensing-execution for once) workflow [17], which may not be optimal for small object bin picking. Most existing approaches first detect the 6D object pose and then plan a proper gripper pose that can pick the target item, followed by executing the plan to pick one item out of the cluster. However, this workflow may suffer from cases when no feasible grasp exists. For example, when small objects are piled closely, a grasp that can perfectly attach the gripper to a target object without collision with the surrounding small objects may not exist.

To this end, this paper proposes a new bin picking framework, called two-stage grasping, aimed at small objects. As illustrated by the name, there are two assorted grasping stages from rough to fine, i.e., sensing-execution for twice. In the "rough" stage, one or more than one items will be grasped out of the bin and placed on a tray. In the "fine" stage, exactly one item will be picked from the tray. Two-stage grasping takes advantage of soft robotic grippers [18-21] and robotic manipulation [22-27] and is compatible with different soft robotic grippers [18-21], existing segmentation methods [2,11,12], pose estimation approaches [8,9,13] and grasp detection methods [16], [28-30]. Besides, by dividing bin picking into two stages, vision detection and object grasping are also divided into rough and fine stages. In either stage, vision detection or object grasping becomes more reliable than in the traditional one-stage workflow. The proposed two-stage bin picking framework provides a promising solution to reliable, robust, accurate, and compatible bin picking for small objects.

The work was supported in part by Hong Kong RGC via the TRS project T42-409/18-R and 14202918, the Hong Kong Centre for Logistics Robotics, CUHK T Stone Robotics Institute, and in part by the CUHK Hong Kong-Shenzhen Innovation and Technology Research Institute (Futian). Corresponding to Jianshu Zhou (Jianshuzhou@cuhk.edu.hk).

HW Cao, JS Zhou, R Cao, JD Huang, and YC Li, and YH Liu are with the T Stone Robotics Institute, the Department of Mechanical and Automation Engineering, the Chinese University of Hong Kong.

Q Dou is with the Department of Compute Science and Engineering, the Chinese University of Hong Kong.

JS Zhou, Q Dou, Ng CM, and YH Liu are also with the Hong Kong Centre for Logistics Robotics (HKCLR).

## II. Small Object Bin Picking System

The hardware system (Fig. 1) contains a robot arm, a soft gripper that performs object singulation and item picking, an RGB-D camera, a movable manipulation tray on which the soft manipulation and visual recognition are conducted, and a placing tray to receive picked items.

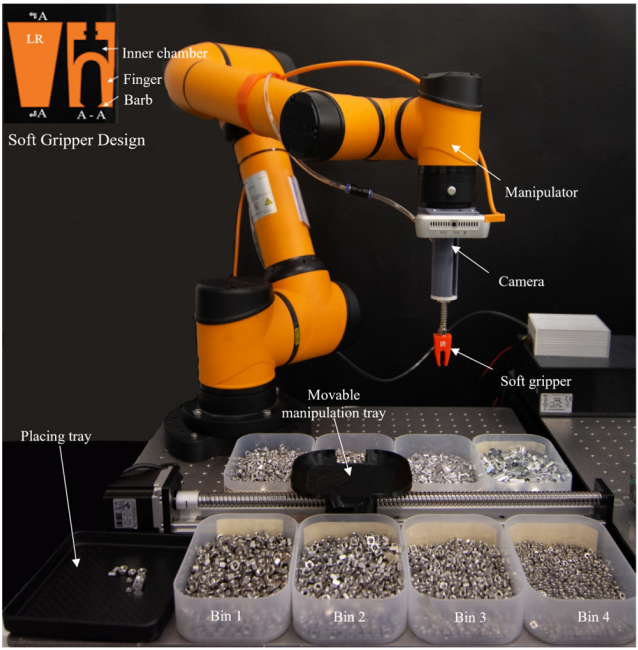

Figure 1. Small object bin picking system composition.

As shown in Fig. 1, the proposed soft gripper contains two soft fingers, a soft palm, and an inner pneumatic chamber [31]. Two fingers close when the inner pressure is positive, i.e., larger than the atmospheric pressure. Conversely, if the inner pressure is negative, i.e., smaller than the atmospheric pressure, two fingers open. By changing the pressure in the inner chamber, the opening width of the soft fingers can be adjusted. Besides, the finger size and action way of the gripper is designed for picking multiple and single small objects.

Two main strengths of the proposed gripper in small object grasping are safety and robustness. The designed gripper is purely soft; therefore, it can absorb extra energy when interacting with small objects, which has two-fold advantages. The first is to decrease undesired impact when grasping. Thus, the gripper can protect objects from damage during grasping and manipulation. The second benefit is that the fingers can adapt to the shape of the small object being grasped, changing point contacts on traditional rigid grippers to the line and surface contacts. Then the possibility of dropping items is decreased. Besides, the fingernails of the soft gripper increase the range of objects it can grasp. We show in experiments that the gripper can pick different shapes and sizes of small items.

## III. TWO-STAGE GRASPING FRAMEWORK

### A. Two-Stage Grasping Overview

The proposed bin picking framework has two stages: a "rough" stage and a "fine" stage. Each stage includes three steps: sensing, planning, and execution, as in the traditional bin picking workflow. In the "rough" stage, rough sensing and planning are conducted in the bin. After executing the plan, roughly some items are grabbed from the bin to a manipulation tray. In the "fine" stage, dedicated sensing and planning are performed on the grabbed items, and precisely one item is picked each time.

The details of two-stage grasping are depicted in Fig. 2. The gripper roughly grabs one or multiple items from a bin at the location predicted by small object density estimation, which will be introduced in Section III-B. The picked objects will then be placed on the manipulation tray for stage two. In the fine stage, the grasp detector looks for feasible grasps on the manipulation tray through dedicated image segmentation and collision checking. Many other grasp detectors could replace this module. If items on the tray are not graspable, the system will try to separate them using soft manipulation. Two singulation policies are presented in this paper: outsweep and break-off. If an item is graspable, the manipulator will execute the planned grasp. Then precisely one item will be picked and placed into the final placing tray. The remaining items left in the manipulation tray are not required and are seen as excess. Thus, they will then be swept back to the bin using soft manipulation called reflow.

One of the advantages of two-stage grasping is that the challenging sensing, planning, and execution of grasps are separated into two stages. In each stage, they become straightforward. No complex grasp detector or pose estimation is needed, though supported.

### B. Density Estimation of Cluttered Small Objects and Rough Grasping

Given an RGB image, this module estimates the density of cluttered identical small objects in the image.

We employ a Fully Convolutional Network (FCN) to predict an object's density map in a pixel-wise manner. Specifically, given an RGB $I \in R^{H \times W \times 3}$, where $H$ and $W$ are image height and width, the density map $p$ is predicted by the FCN $\mathcal{F}$ as:

$$p = \mathcal{F}(I) \quad (1)$$

In our implementation, we use U-Net [32] backbone to predict density values for every pixel. For training, we directly regress the density map using mean squared error (MSE) loss as:

$$\ell(p, \bar{p}) = \frac{1}{H}\frac{1}{W}\sum_{i=1}^{H}\sum_{j=1}^{W}\left(p(i,j) - \bar{p}(i,j)\right)^2 \quad (2)$$

, where $i$, $j$ demote pixel location, and $p$, $\bar{p}$ represent the predicted and ground truth density map, respectively. To avoid overfitting during training, we use data augmentation (e.g., random Gaussian noise and affine transformation) on RGB images following [33].

Due to the large number of small objects in the scene, manual labelling is difficult and time-consuming to get a precise ground truth density map. To bypass a real data collection procedure, we generate photo-realistic synthetic (PBR) data for training based on our PBR image generation pipeline [8]. Specifically, we randomize i) environment light, ii) camera location, iii) objects' pose, and iv) background to narrow the domain gap between real and synthetic data, as shown in Fig. 3. After obtaining the RGB image with the objects' pose in a virtual scene, we can represent objects as dots projected in the image plane (see Fig. 3(b)). Then, the ground truth density maps can be generated by applying the gaussian kernel over the dot map (see Fig. 3(c)). We collect six industrial objects' CAD models for the training set. One hundred synthetic scenes of cluttered objects are rendered for each object based on Blender. The number of objects for each scene varies from 100 to 300. We manually label dot maps for testing data and use the same protocol to process the testing set. We empirically found that the model trained by synthetic data is already sufficient to predict precise density maps for downstream grasping tasks.

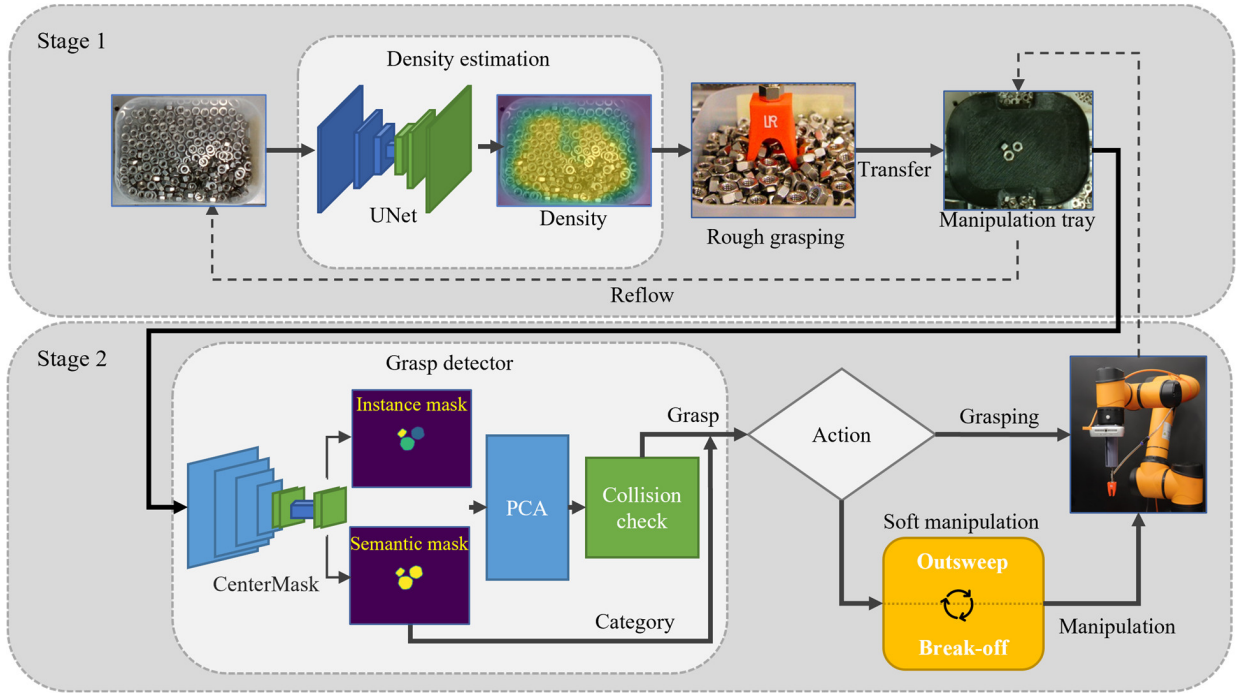

Figure 2: Two-stage grasping framework. The proposed bin picking framework has two stages: a "rough" stage and a "fine" stage. In stage 1, the gripper roughly grabs some items from the bin at the location instructed by density estimation. The grabbed items are placed on a manipulation tray. In stage 2, a grasp detector makes decisions based on images of the manipulation tray to pick or separate objects. If no feasible grasp exists, soft manipulation will be conducted on the manipulation tray to separate items there. Each time exactly one item will be picked by the gripper from the tray. Finally remaining items will reflow to the bin via soft manipulation.

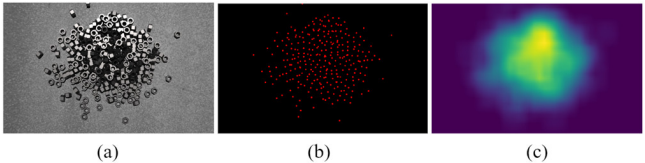

Figure 3. Example of a synthetic RGB image (a), the corresponding dot map (b), and the density map (c).

*C. Grasp Detection*

A deep learning-based grasp detection module is introduced. It is noted that two-stage grasping is compatible with different segmentation methods [2,11,12], pose estimation approaches [8,9,13] and grasp detection methods [16], [28-30].

We use a real-time anchor-free segmentation package, CenterMask [34]. The model is trained on a self-collected dataset. The segmentation module outputs an instance mask based on which grasps are generated and a semantic mask that indicates the category of each item.

The segmentation results are passed to a 2D grasp detector. The grasp detector first performs contour finding and PCA [35] on the segmented image of the scene so that the planar orientation $\vec{p}$ of each item can be estimated. Secondly, contacts can be generated based on the orientation and contour of each item. Since the soft gripper has two fingers, a grasp $G(\zeta, \theta)$ is defined by two contact points $s_1\ (u_1, v_1)$ and $s_2\ (u_2, v_2)$ on the boundary of an item in the image, with the constraint that $\vec{p} \cdot \overrightarrow{s_1 s_2} = 0$:

$$\zeta = \left(\frac{u_1 + u_2}{2}, \frac{v_1 + v_2}{2}\right) \quad (3)$$

$$\theta = \cos^{-1} \frac{\overrightarrow{s_1 s_2} \cdot \vec{\iota}}{|\overrightarrow{s_1 s_2}|} \quad (4)$$

, where $\vec{\iota}$ is the unit vector in the direction of the x-axis in the image plane. Thirdly, after generating contacts and grasps, collision checking is conducted by placing virtual fingerprints at every contact point and performing pixel-wise clearance checking. Only grasps corresponding to isolated objects can survive. Finally, all collision-free and motion-feasible grasps and categories are passed to the task planner. To execute grasps, actuation pressure is mapped linearly from $\|\overrightarrow{s_1 s_2}\|$.

*D. Small Object Singulation by Soft Manipulation*

Object grasping has proven reliable given objects in isolation on a flat surface. Some grasping methods have been extended to clustered objects. However, precise segmentation and localization of small objects, such as nuts and bolts, in clusters remains difficult. For example, a 5 mm localization error could result in enclosing more than one item into the gripper or enclosing nothing, which is not desired. Singulated objects are more favorable than cluttered objects because they can be grasped by 2D grasp methods with faster computation and higher reliability. In case objects to be grasped are piled up, except for using grasping methods extended to clustered objects, one strategy is to separate them in isolation first and then grasp each of the items easily.

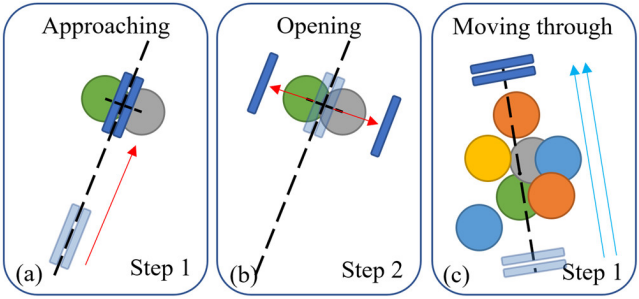

Figure 4: Small object singulation policies. Color circles denote objects. The blue rectangles and red arrows show the motion of the soft gripper and movement of fingers. (a)-(b) Outsweep policy; (c) Break-off policy.

Pushing actions can potentially separate cluttered objects [22-27], i.e., separate one item away from others, creating more reliable and collision-free 2D grasps. In this paper, we proposed two pushing-like small item singulation policies (Fig. 4.), outsweep policy and break-off policy with the soft gripper and a soft manipulation algorithm.

Unlike traditional pushing policies that only contain trajectories of grippers, outsweep policy uses the degrees of freedom of the soft fingers. It generates pushing motion using the following steps:

1) Find two closest items in the cluster whose centers are $C_1$ and $C_2$ and the midpoint $C_0 = (C_1 + C_2)/2$,

2) Construct a line $\overline{C_1 C_2}$ and a vector $\overrightarrow{P_i C_0}$ that is perpendicular to $\overline{C_1 C_2}$, where $P_i$ is an accessible space point,

3) Generate a motion sequence that includes approaching $C_0$ along $\overrightarrow{P_i C_0}$ with gripper orientation aligning to $\overrightarrow{P_i C_0}$, stopping at $C_0$, and finally opening fingers rapidly at $C_0$.

Break-off policy generates pushing motion using the following steps:

1) Find the center of the cluster $C_i$ and a line passing through $C_i$ that starts and ends at the free space point $P_i$ and $P_j$,

2) Construct two vectors $\overrightarrow{P_i P_j}$ and $\overrightarrow{P_j P_i}$ with opposite directions, and a vector $\overrightarrow{P_i P_j}_\perp$ that perpendicular to the vectors in the plane,

3) Generate two pushing motions, one for each vector, that guide the gripper moving along the vector while keeping the orientation aligned to $\overrightarrow{P_i P_j}_\perp$. Choose either motion.

Then one following question is how to choose one from the above two policies to execute each time. In this paper, a heuristic soft manipulation algorithm is described in Alg. 1. The algorithm takes the size of a cluster and outputs the proper singulation policy to be executed on that cluster. Here we use k-means [36] on coordinates of item centers to cluster them. Experiments have shown the effectiveness of efficiency of the singulation manipulation algorithm. More singulation policies will be proposed in our future work.

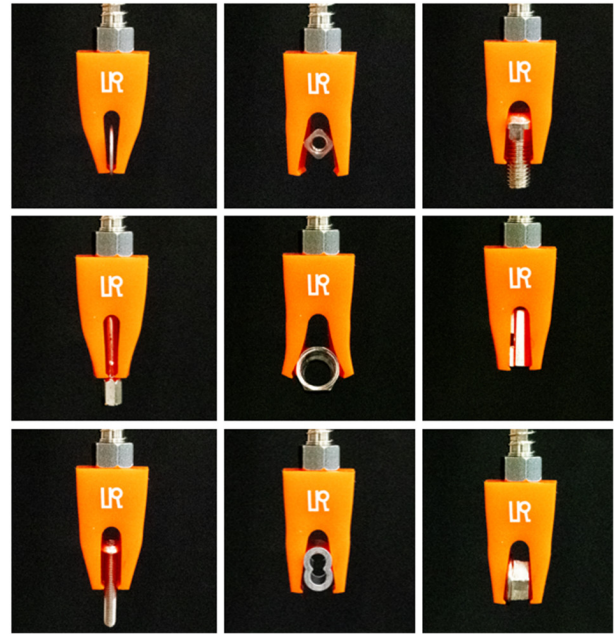

Figure 5: Small object grasping experiment. Objects with different sizes and shapes are grasped by the soft gripper.

**Algorithm 1:** Soft Singulation Manipulation

    **Input**: Instance segmentation of the scene
    **Output**: The policy to be executed
    *threshold* = 3;
    Find clusters $\Phi$ of items using k-means clustering;
    Randomly sample a cluster $\varphi$ out of $\Phi$;
    Compute the size $s$ of $\varphi$ as the number of items;
    **if** $s \leq threshold$ **then**
        **return** *outsweep_flag*, $\varphi$;
    **else**
        **return** *break-off_flag*, $\varphi$;

## IV. EXPERIMENTAL VALIDATIONS

Several experiments were carried out to evaluate the performance of the proposed bin picking framework, the soft gripper, and the soft manipulation method. Firstly, we evaluated the grasping capability of the soft gripper. Then, the soft manipulation performance was measured. Thirdly, a realization of the bin picking framework was presented. The grasp success rates of two-stage grasping and of the traditional bin picking framework were compared. RealSense D415 camera was used in experiments.

### A. Small Object Grasping by the Soft Gripper

The soft gripper's grasping capability is essential for picking, sorting, and dispensing small items. In our experiments, we collected multiple industrial objects and mechanical fasteners of different shapes and weights and controlled the gripper to grasp them. The widths of objects range from 1.2 mm to 25 mm. As shown in Fig. 5, all of them can be successfully grasped by the soft gripper. Moreover, as described in Section IV-D, the gripper can open its fingers wider to grab more than one item each time.

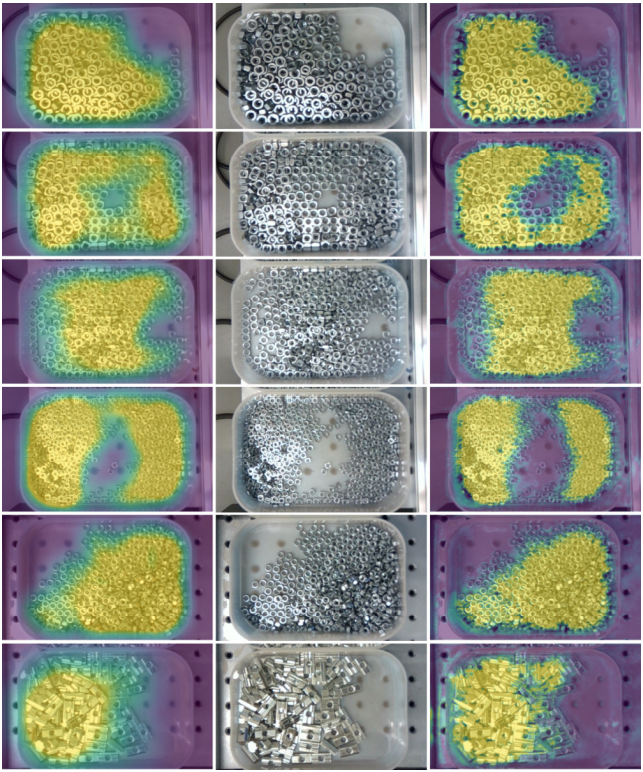

Figure 6: Visualization of example results on the test images. From left to right of each row: ground truth, real image, and the predicted density.

### B. Density estimation for rough grasping

To test the density estimation model's generalization capacity, we trained it using synthetic data and directly tested it on real data without fine-tuning. The testing results are presented in Table I, and examples of predicted density map and tested objects are shown in Fig. 6. Noticed that the object shapes in the real scenario is not same with the ones in the training set, showing a good generalization towards unseen objects. The results show consistent effectiveness towards different objects in different scenes.

TABLE I. MEAN SQUARED ERRORS FOR DIFFERENT OBJECTS ON THE REAL TESTING SET

| Object | 1 | 2 | 3 | 4 | 5 | 6 | 7 | 8 |
|---|---|---|---|---|---|---|---|---|
| Errors | 0.08 | 0.12 | 0.09 | 0.09 | 0.092 | 0.10 | 0.14 | 0.19 |

### C. Soft Manipulation for Small Item Singulation

We compared our methods with a baseline singulation policy to test the effectiveness of our proposed singulation policies and the soft manipulation algorithm. The baseline adopted in this experiment was modified from [23]. Specifically, the baseline policy is to push a cluster of items along a trajectory, passing through any point in the cluster with a random starting point and pushing distance. Our methods to be compared against the baseline and each other includes outsweep policy, break-off policy, and soft manipulation algorithm that automatically chooses between outsweep and break-off. Some small objects were randomly chosen as the testing objects. In each run, the bin picking system was tasked with picking a cluster of small objects. If no feasible grasp exists, singulation manipulation was conducted. Generally, clusters that contain more items may need more times of singulation manipulation to complete picking. Each singulation method was conducted on six kinds of scenarios: clusters of two, three, four, six, ten, and twenty items. The items cluster was randomly formulated. The items were randomly chosen. Five trials were performed in each scenario. In total, 120 trials were performed. In each trial, the times of singulation manipulation (picking actions were not counted) were recorded and plotted in Fig. 7.

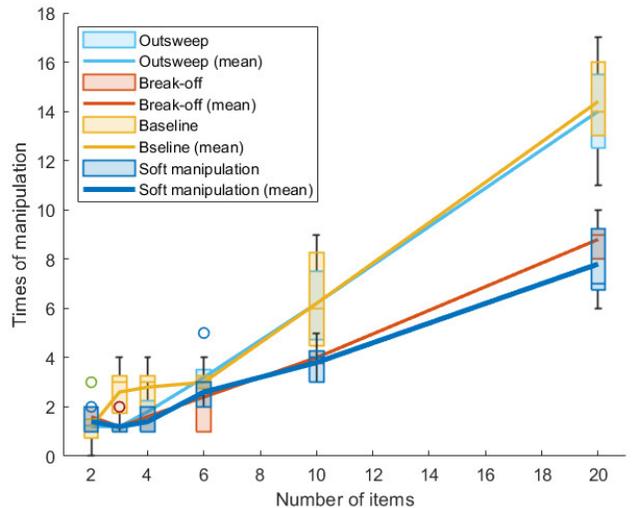

Figure 7: Small object singulation experiment. Different policies were compared.

When the cluster is small, such as two or three items, outsweep and soft manipulation algorithm could finish picking with less singulation manipulation. If the cluster is large, such as more than four objects, break-off and soft manipulation algorithm could complete picking with less singulation manipulation. In all cases, baseline needed more times of manipulation, implying that it was the most time-consuming. These results indicated that the soft manipulation algorithm is suitable for efficiently separating multiple small objects under different circumstances.

### D. Realization and evaluation of Two-Stage Grasping

Manipulators accurately grasping a small item with a random pose out from a bin is challenging due to factors like undesired object surface and sizes, pose estimation uncertainty, and sensor resolution limitations, as mentioned in Section I. To this end, two-stage grasping was implemented with the bin picking system, and snippets were shown in Fig.8.

The task was to pick the desired number of items from each bin using two-stage grasping. In Fig. 8 (a1)-(a7), the experimental process regarding the second bin was taken as the first example. The system estimated object density to locate the rough grabbing area in the bin, as shown in Fig. 8 (a1). A rough grasping was performed, and several picked items were placed onto the manipulation tray, as shown in Fig. 8 (a2) and (a3). The grasp detector decided to precisely pick the desired number (one in this experiment) of items one by one on the manipulation tray, as shown in Fig. 8 (a5). After fine picking, a sweeping action returned the remaining items to the bin, as shown in Fig. 8 (a7). Fig. 8 (b1)-(b7) showed another example. In the previous example, items were fortunately isolated upon placing them in the manipulation tray. Hence no singulation manipulation was conducted. Sometimes, singulation manipulation is necessary, as shown in Fig. 8 (b5). In this example, as shown in Fig. 8 (b4), the tray items were closely near each other, so the grasp detector stated them as infeasible to grasp. The result of outsweep singulation was shown in Fig. 8 (b5). Two feasible grasps were newly detected after singulation. One of them was picked, and the other was swept back to the bin, as shown in Fig. 8 (b6). The remaining process was similar and not shown.

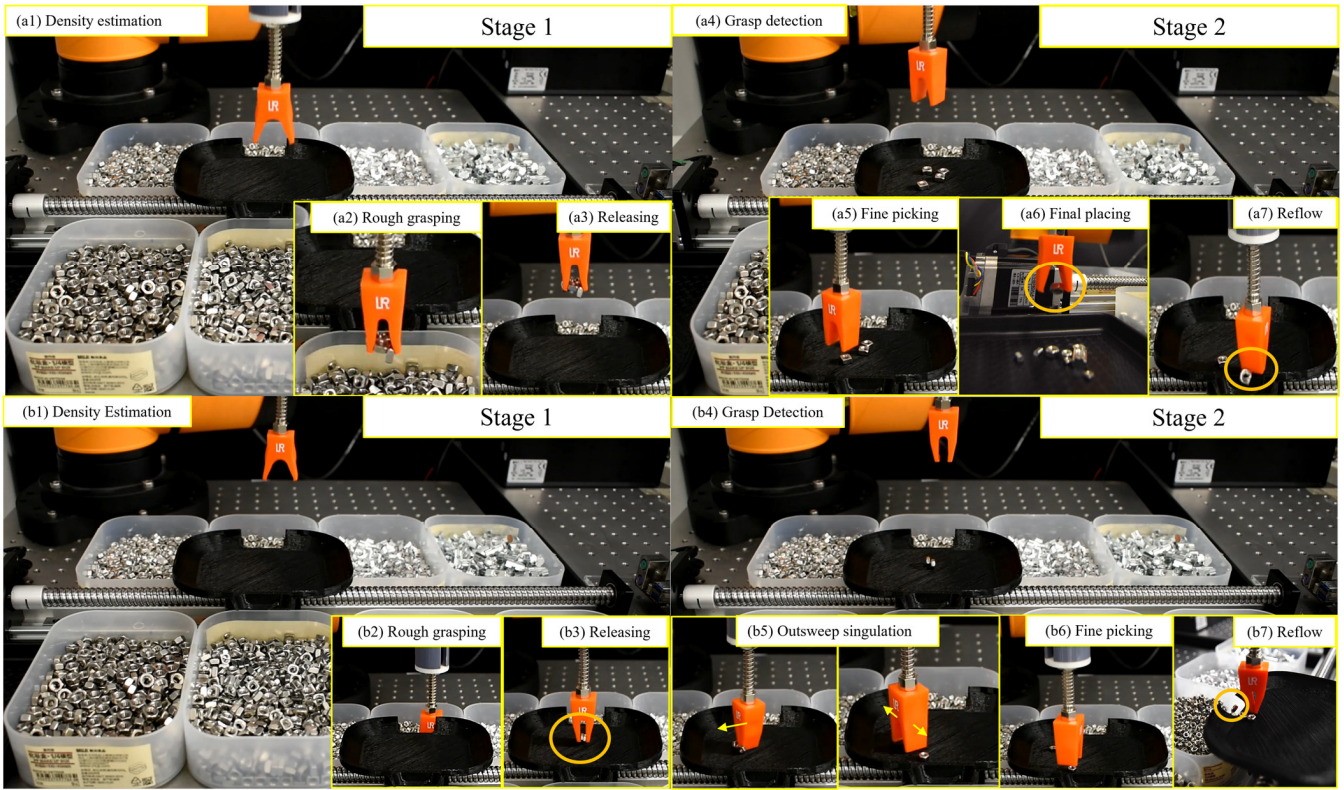

Figure 8: Automatic bin picking process using two-stage grasping with the small object bin picking system. Two examples are presented in (a1-a6) and (b1-b6), respectively.

We compared the proposed bin picking framework with the traditional one-stage workflow under the same setting. Each framework was tasked with picking one small object from every bin. The experiment was repeated five times. The manipulator speed was set to 30% of the maximum. Results are shown in Table II. Compared to traditional one-stage workflow, two-stage grasping achieved much a better success rate but needed longer execution time. The time is mostly spent on robot arm movements, such as item transferring to the moveable and placing tray.

TABLE II.    PICKING SUCCESS RATES AND TIME COSTS

|                      | Two-Stage Grasping | Traditional Bin Picking Pipeline |
|----------------------|--------------------|----------------------------------|
| **Success Rate**     | 92.5%              | 37.5%                            |
| **Time Cost per Cycle** | 50.1 seconds    | 29.9 seconds                     |

## V. CONCLUSION AND FUTURE WORK

We proposed a novel bin picking framework for small objects in this work. The proposed two-stage grasping framework differs from traditional bin picking pipelines in sensing, planning, and executing. By separating them into the fine and rough stages, challenges in small object bin picking resulting from sensor resolution limitations, image or point cloud processing difficulties, and collision detection are simplified. A small object density estimation method has been proposed for rough grasping. Soft manipulation algorithm has been introduced for isolating cluttered objects. Two small object singulation policies have been presented. Experiments have shown that the proposed two-stage grasping, soft gripper, soft manipulation algorithm, and density estimation method have achieved remarkable results.

In future work, we will refine the soft robotic gripper by integrating soft sensors to its body and surface for proprioception and haptic realization. Also, we plan to design a novel actuation method for this soft gripper to get rid of the commercially available but bulky pneumatic actuation. Furthermore, a data-driven model with deep reinforcement learning techniques will improve the soft manipulation algorithm.